\documentclass[conference]{IEEEtran}
\IEEEoverridecommandlockouts
\usepackage{cite}
\usepackage{amsmath,amssymb,amsfonts}
\usepackage{algorithmic}
\usepackage{graphicx}
\usepackage{textcomp}
\usepackage{xcolor}
\usepackage{booktabs}
\usepackage{tabularx}
\usepackage{placeins} 
\def\BibTeX{{\rm B\kern-.05em{\sc i\kern-.025em b}\kern-.08em
    T\kern-.1667em\lower.7ex\hbox{E}\kern-.125emX}}
\begin{document}

\title{Multi-Task Self-Supervised Learning for Image Segmentation Task\\}

\author{
\IEEEauthorblockN{Lichun Gao} 
\IEEEauthorblockA{\textit{Computer Science Department} \\
\textit{Worcester Polytechnic Institute}\\
Worcester, MA \\
lgao2@wpi.edu}
\and
\IEEEauthorblockN{Chinmaya Khamesra}
\IEEEauthorblockA{\textit{Robotics Department} \\
\textit{Worcester Polytechnic Institute}\\
Worcester, MA \\
ckhamesra@wpi.edu}
\and
\IEEEauthorblockN{Uday Kumbhar}
\IEEEauthorblockA{\textit{Data Science Department} \\
\textit{Worcester Polytechnic Institute}\\
Worcester, MA \\
ukumbhar@wpi.edu}
\and
\IEEEauthorblockN{Ashay Aglawe}
\IEEEauthorblockA{\textit{Data Science Department} \\
\textit{Worcester Polytechnic Institute}\\
Worcester, MA \\
alaglawe@wpi.edu}

}

\maketitle

\begin{abstract}
Thanks to breakthroughs in AI and Deep learning methodology, Computer vision techniques are rapidly improving. Most computer vision applications require sophisticated image segmentation to comprehend what is image and to make an analysis of each section easier. Training deep learning networks for semantic segmentation required a large amount of annotated data, which presents a major challenge in practice as it is expensive and labor-intensive to produce such data. The
paper presents 1. Self-supervised techniques to boost semantic segmentation performance using multi-task learning with Depth prediction and Surface Normalization . 2. Performance evaluation of the different types of weighing techniques (UW, Nash-MTL) used for Multi-task learning. NY2D dataset was used for performance evaluation. According to our evaluation, the Nash-MTL method outperforms single task learning(Semantic Segmentation).    
\end{abstract}

\section{Introduction}
Deep learning is now recognized as a standard strategy for problems such as classification, segmentation, and detection, as computer vision and machine learning have improved rapidly. A vast number of cutting-edge techniques rely on supervised learning, which necessitates manual data labeling, which is both time-consuming and costly.

Unlabeled images and videos can be found in big quantities for a low price. Sadly, their full potential is rarely realized. Unsupervised learning is used for uncovering hidden patterns in unlabeled data but isn't designed to solve a specific problem, therefore it misses out on important information needed to complete visual tasks (e.g., segmentation).

Self-supervised learning has the potential to overcome limitations and capitalize on the advantages of both supervised and unsupervised learning. It is a sort of supervised learning in which labels are generated automatically from unlabeled data. As a result, unlike unsupervised learning, self-supervised learning concentrates on optimizing a specific task, forcing the network to acquire semantic knowledge without having to deal with additional label-related concerns.

The majority of previous supervised and self-supervised learning research has focused on a single task at a time. This yields good results, but it overlooks a lot of relevant data. When many tasks are simultaneously trained, domain-specific knowledge is utilised to a greater extent, resulting in better generalization.\\

\subsection{Research Contribution}
The previous work was done on weighting techniques like grid, search, uncertainty weighing (UW) and Dyanmic Weigh average. From the previous works, it was concluded that uncertainty weight method (UW) performed better than Dynamic weight average (DWA). As an extension to the work we tried the Nash-MTL weight method which gave us the best results when compared to UW weighting method.

\section{Related Works}
The two self-supervised tasks that we are focusing on are Surface Normal prediction and Depth prediction. We fulfill the tasks for semantic segmentation based on the field of deep learning. Because of multiple tasks included, we apply multi-task learning to improve the efficiency of our experiment. We will talk about the details in the following sections. 

\subsection{Self-Supervised Learning}
Self-supervised learning is a useful tool for us to learn magnitude of more data, which helps the large-scale model to be trained even without labels.  Self-supervised learning learns the signal from the original data, the goal of self-supervised learning is to predict the hidden part or the part which has not been observed for the input value in the context of un-hidden parts. The important thing is that self-supervised learning will generate ground truth labels automatically. Our solution for the tasks 1) surface normal prediction and 2) depth prediction are using self-supervised learning.The limit for self-supervised learning is mostly about the under performance compared with supervised learning. However, this problem does not influence our experiment based on our tasks.\\

\subsubsection{Surface Normal Prediction}
Surface Normal prediction is the task of predicting the surface orientation of the objects present inside a scene. There are many researchers who have worked in this area, and they have found many methods to accomplish this task. Eigen et al. used surface normals in their convolutional architecture study. Qi et al. also applied to their study about joint depth prediction and surface normal estimation. From the previous study, we could also tell that surface normal prediction usually comes with the task of depth prediction. Considering this part, we include both tasks in our study.\\
For surface normal prediction, we will have the following formula:\\ \hspace*{\fill} \\
\centerline{$L(I,Y)=-\sum_{i=1}^{M\times M}\sum_{k=1}^{K}(\Bbb1 (y_{i}=k)logF_{i,k}(I))$}\\ \hspace*{\fill} \\
Where $F_{i,k}(I)$ is the probability for $i$th pixel, which should have the normal defined by the $k$th code. The $\Bbb1 (y_{i}=k)$ represents the indicator function, and $Y={y_{i}}$ is the set of ground truth labels for surface normal prediction. Moreover, $M=M_{t}$ and $K=K_{t}$.\\

\subsubsection{Depth Prediction}
Predicting the depth is an indispensable task for understanding the 3d scene. The task is ambiguous to some extent because of the overall scale. Depth prediction takes an important role in autonomous driving, that is the reason why many researchers are focusing on this area right now. Depth prediction in supervised learning had shown promising results, however, the cost is huge because we need to utilize the sensor and human resources to do this task. To make this task cost-effective, the research goal has shifted to self-supervised learning and un-supervised learning. Based on the work from Godard et al. we start to learn the depth by capturing the images from different viewpoints with simple geometric equations. With the improvement, Sudeep et al. discovered the bottleneck to the depth prediction performance is the low image resolution. Consequently, Vitor et al. introduced us to a neural network architecture specialized in self-supervised monocular depth estimation. With the development of technology, we then have many helpful datasets. For instance, the benchmark Dense Depth for Automated Driving (DDAD). The results have been improved for much following work.\\
In our training, we will apply both left ($I^{l}$) and right ($I^{r}$) images to predict, transforming the right image into the left image. We will reconstruct the left image based on the following formula, where $\hat{I} ^{l}$ denotes our reconstructed left image, and bilinear denotes the bilinear interpolation function:\\ \hspace*{\fill} \\
\centerline{$\hat{I} ^{l}$ = $bilinear$($I^{r},d$)}\\ \hspace*{\fill} \\
After this, we will obtain a reconstruction loss similar as the $L2$:\\
\centerline{$L_{depth} (I^{r},I^{l},\hat{I} ^{l}) = \frac{1}{N} \sum_{i=0}^{N-1}(I_{i}^{l}-\hat{I}_{i}^{l})^{2}$}\\ \hspace*{\fill} \\
We will use sigmoid function to convert the output of the final convolutional layer to disparity. Finally, we could obtain the depth from the disparity.

\subsection{Semantic Segmentation}
Semantic segmentation is one of the tasks that researchers focus on most in the field of computer vision, it could be used for three steps: 1) classifying, 2) localizing, and 3) segmentation. To be more specific, Semantic segmentation is the process to classify pixels and generate a label for each pixel. Semantic segmentation is important because it can derive correlation of the input image, and remove the noise. In deep learning methods, the convolutional neural network is frequently used to perform this task. Long et al. had brought the first fully convolutional network for semantic segmentation to our view. This was a milestone because it highly improves efficiency and accuracy. Vijay et al. had proposed the SegNet, which is based on CNN. Other solutions include UNet, PSPNet, PANet, and DANet. However, with the limitation of computational requirements, many of those networks do not fit with the industry. To address the issues in the industry, smaller networks such as ENet, MobileNet, et al. are proposed.\\
The formula of cross-entropy loss for the semantic segmentation is:\\ \hspace*{\fill} \\
\centerline{$L_{seg}(S,\hat{S}) = -\frac{1}{N}\sum_{i=0}^{N-1}S_{i}log(\hat{S}_{i})$}\\ \hspace*{\fill} \\
Since ground truth label is required in semantic segmentation, $S_{i}$ is the ground truth for i-th pixel. $\hat{S}_{i}$ denotes the class prediction that:\\ \hspace*{\fill} \\
\centerline{$\hat{S_{i}} = e^{zs_{i}}/{\textstyle \sum_{s}^{}e^{zs_{i,s}}} $}\\\hspace*{\fill} \\ 
Where $zs$ is the output of the final convolutional layer of the decoder. $s$ is the number of semantic classes.  

\subsection{Multi-task Learning}
Multi-task learning is an important learning paradigm whose goal is to leverage the essential information from multiple relative tasks. The main goal for multi-task learning is to improve the general performance among all tasks. In the image field, Marvin et al. proposed Multinet, the first architecture for classification, detection, and segmentation. It includes the encoder (VGG) and the decoder (classification decoder and segmentation decoder).

Among all of the developed multi-task learning models, the one that fits our work mostly is the multi-task self-supervised visual learning raised by Carl et al. It has trained different kinds of complementary self-supervised tasks simultaneously to obtain the best learning results. Those tasks include: 1) predicting relative position, 2) color prediction, 3) single-sample learning, and 4) motion segmentation. Two structures are used in their work, making the network more flexible. This work is engineering-oriented but provides us with significant research ideas. 

Because we are using self-supervised tasks, the loss for our work the weighted sum of task-specific losses: \\ \hspace*{\fill} \\
\centerline{$L_{total}=\lambda_{1}L_{surface}+\lambda_{2}L_{depth}+\lambda_{3}L_{seg}$}\\ \hspace*{\fill} \\
Where $\lambda_{1},\lambda_{2},\lambda_{3}$ represents the weight for 1)surface normal prediction, 2)depth prediction, 3)semantic segmentation.  

\section{Proposed Method}
\subsection{Nash-MTL}
In multi-task learning (MTL), a joint model is trained to make predictions for multiple tasks at the same time. Joint training saves time and money by reducing computation costs and increasing data efficiency; however, because the gradients of these different tasks can conflict, training a joint model for MTL often results in lower performance than its single-task counterparts.

A common solution to this problem is to combine per-task gradients into a joint update direction using a specific heuristic.
Here, we propose viewing the gradients combination step as a bargaining game in which tasks negotiate to reach an agreement on a joint parameter update direction. Under certain assumptions, the bargaining problem has a unique solution known as the Nash Bargaining Solution, which we propose to use as a principled approach to multi-task learning.

Based on Nash's findings, we propose Nash-MTL, a novel MTL optimization algorithm in which the gradients are combined at each step using the Nash bargaining solution.
We begin by characterizing the Nash bargaining solution for MTL and developing an efficient algorithm for approximating its value. Then, in the convex and nonconvex cases, we theoretically analyze our approach and establish convergence guarantees. Finally, we demonstrate empirically that NashMTL approach achieves cutting-edge results across a range of challenges.

\subsection{Uncertainty Weighting}
Multi-task learning is concerned with the problem of optimizing a model in terms of multiple objectives. To combine multi-objective losses, the approach would be to simply perform a weighted linear sum of the losses for each individual task.

This method, however, has a number of drawbacks. Model performance, in particular, is extremely sensitive to weight selection. These weight hyper-parameters are costly to tune, often taking several days per trial. As a result, it is preferable to find a more convenient approach capable of learning the optimal weights.

Consider a network that learns from an input image to predict pixel-wise depth and semantic class. We find that at some optimal weighting, the joint network outperforms separate networks trained on each task separately. The network performs worse on one of the tasks when the weights are close to the optimal value. However, finding these optimal weightings is costly and becoming increasingly difficult with larger models.

Task uncertainty captures the relative confidence between tasks, reflecting the uncertainty inherent in the regression or classification task. It  also depends on how the task is represented or measured. We propose that in a multi-task learning problem, we can use homoscedastic uncertainty as a basis for weighting losses.

\section{EXPERIMENTS}
\subsection{Dataset} 
We used NYU-Depth V2 data set which is comprised of video sequence from a variety of indoor scenes.The dataset consist of labelled and raw images.  It consist of 1449 densely labeled pairs of aligned RGB and dept images and 407,024 labelled frames. 

\subsection{Architecture}
The model was built using PyTorch. The architecture consists of two steps: 
Spatial Pooling Pyramid and Encoder-Decoder Network. The Spatial Pooling Pyramid captures the images at multiple scale which is required for computer vision tasks. The network architecture uses U-net, in which outputs of the encoder layer are combined with the inputs of the decoder layers through concatenation. Resnet is set as the backbone of the model. Batch normalization with ReLU is used. The last layer is followed by task-specific activation functions.

 \begin{figure}[htp]
    \centering
    \includegraphics[scale=0.3]{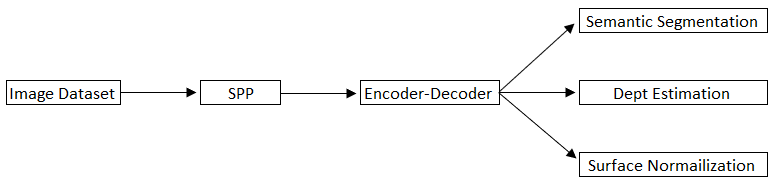} 
    \caption{Model Architecture}
    \label{fig:fashion}
\end{figure}

The model was trained for 200 epochs/iterations. FIrst, the model was trained using only Single task i.e. Segnet. Then the model was trained using Multi-task learning using depth prediction and Surface Normal prediction using weighting methods like Nash-MTL and UW(Uncertainty Weighing). Delta M was calculated to compare the baseline model(Segnet) with other multi-task methods with different weighing approaches.

\section {Results}
  \begin{figure}[htp]
    \centering
    \includegraphics[scale=0.7]{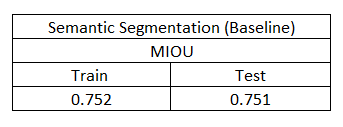} 
    \caption{Semantic Segmentation(Baseline)}
    \label{fig:fashion}
\end{figure}
In Fig2, the MIOU for semantic segmentaiton (Segnet) for test and train is calculated. 
 \begin{figure}[htp]
    \centering
    \includegraphics[scale=0.8]{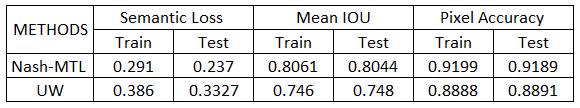} 
    \caption{Semantic Segmentation}
    \label{fig:fashion}
\end{figure}

\begin{figure}[htp]
    \centering
    \includegraphics[scale=0.8]{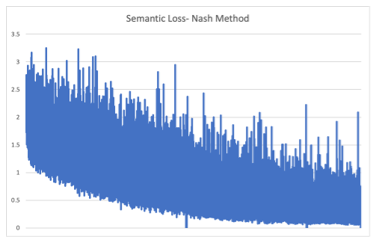} 
    \caption{Semantic Loss: NASH-MTL 200 epochs}
    \label{fig:fashion}
\end{figure}

In Fig(3), the Semantic loss was evaluated for Nash-MTL and UW techniques and Nash-MTL has less Semantic loss or More MIOU when performed on the test dataset. Fig(4) plots the Semantic loss per epoch. 

 \begin{figure}[htp]
    \centering
    \includegraphics[scale=0.7]{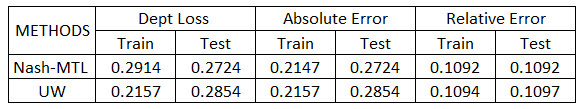} 
    \caption{Dept Estimation}
    \label{fig:fashion}
\end{figure}

\begin{figure}[htp]
    \centering
    \includegraphics[scale=0.5]{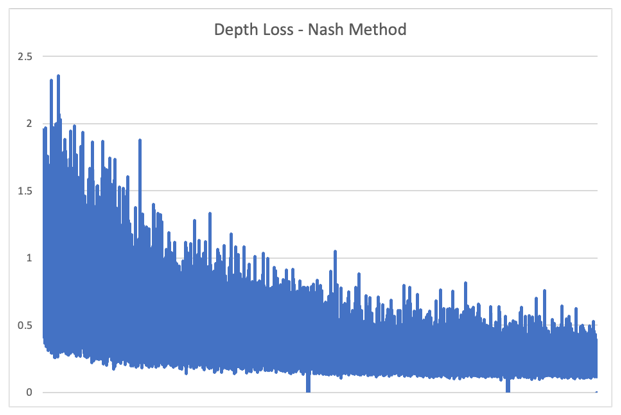} 
    \caption{Depth Loss: Nash-MTL 200 epochs}
    \label{fig:fashion}
\end{figure}

In Fig(5), the Depth loss was evaluated for Nash-MTL and UW techniques and Nash-MTL has less depth loss when performed on the test dataset. Fig(4) plots the Depth loss per epoch. 

 \begin{figure}[htp]
    \centering
    \includegraphics[scale=0.45]{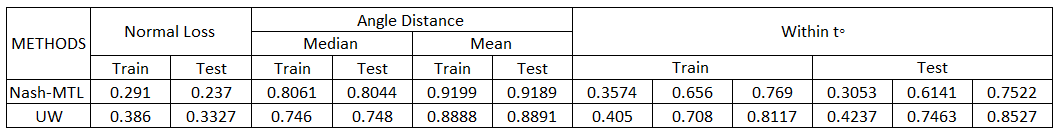} 
    \caption{Surface Normal}
    \label{fig:fashion}
\end{figure}

\begin{figure}[htp]
    \centering
    \includegraphics[scale=0.8]{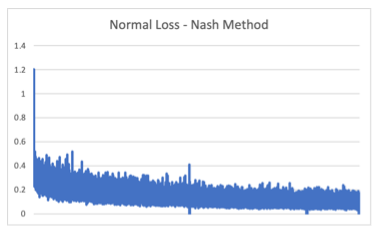} 
    \caption{Surface Normal Loss: Nash-MTL 200 epochs}
    \label{fig:fashion}
\end{figure}

In Fig(7), the Normal loss was evaluated for Nash-MTL and UW techniques. Nash-MTL has less normal loss when performed on the test dataset. Fig(8) plots the Normal loss per epoch. 

 \begin{figure}[htp]
    \centering
    \includegraphics[scale=1]{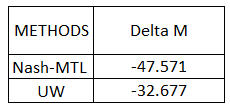} 
    \caption{Delta M}
    \label{fig:fashion}
\end{figure}

In fig(9), Delta M is used as a metric to evaluate weighting techniques in comparison to the baseline. Nash MTL has smaller Delta-M in comparison to UW when Segnet is kept as the baseline for the comparison.
Fig(10) visualises the semantic segmentation and depth estimation of a RGB picture. 

 \begin{figure}[htp]
    \centering
    \includegraphics[scale=0.3]{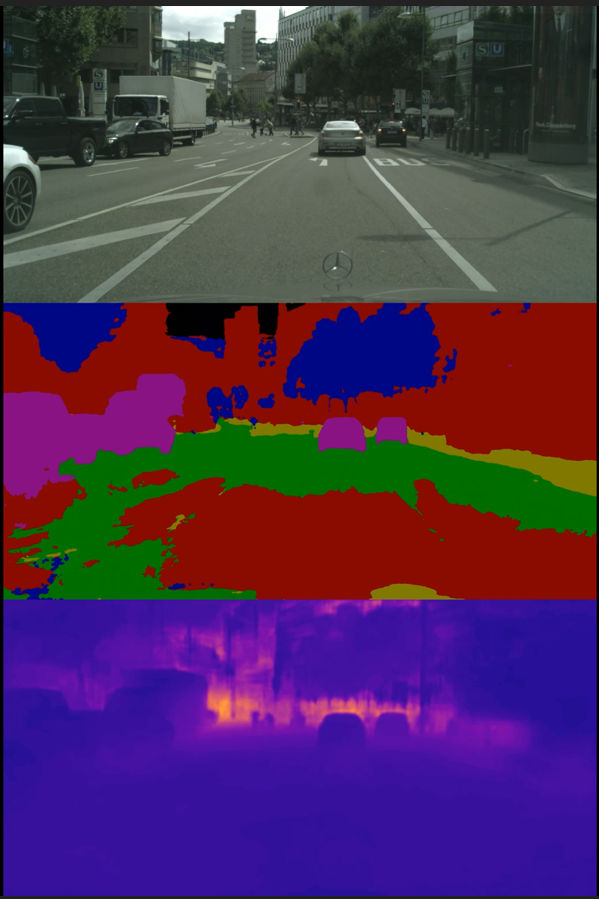} 
    \caption{Visualisation}
    \label{fig:fashion}
\end{figure}

\section{Discussions}
Multiple tasks are addressed at the same time in multi-task learning, usually with a single neural network. MTL has advantages such as improved pixel precision, higher mean IOU, and lower training loss, but it comes at the cost of increased model inference time. In MTL, using Nash weighting method outperforms most of the new published task weighting methods. Using combination of Depth Estimation and Semantic Segmentation as a tasks can yield better results.\\
In our work, we mainly focus on semantic segmentation, surface normal prediction, and depth prediction. We may have other combination of tasks which will may more effectively fulfill object-oriented tasks.\\
Furthermore, the functions and models we used are the most suitable ones for our work, but not the best ones considering time and running issues. And we don't have our own GPU so we used Turing to run our experiment. This caused a limitation to the scale of data we could select, which may have an influence on our accuracy and loss.\\

\section{Conclusions and Future Work}

We used multi-task learning paradigm to boost multiple self-supervised tasks in the field of computer vision. We have learned several things: 1)self-supervised learning is significant for computer vision. Without it, the cost of labeling will be huge. 2)multi-task learning is a useful paradigm to improve task efficiency generally.Multi-task when used with Nash-MTL weighing method outperforms single task semantic segmentation(Segnet).   \\
Better results could be obtained with use of data augmentation. Experimenting with different weighting methods and finding optimal inference time. The work could be extended to new semantic segmentation dataset like waymo(opendatasetv130: perception dataset).

\end{document}